\documentclass[journal]{IEEEtran}
\usepackage{cite}
\usepackage{booktabs}
\usepackage{arydshln} 
\usepackage{multirow}
\usepackage[table]{xcolor}
\usepackage{amsmath}
\usepackage{multirow}
\usepackage{algorithm}
\usepackage{algorithmic}
  
\usepackage{array}

\usepackage[colorlinks,
linkcolor=green, 
anchorcolor=blue,
citecolor=blue]{hyperref}

\usepackage{color}            

\usepackage{mathrsfs}
\usepackage{graphicx}
\usepackage{subfigure}
\usepackage{float}
\usepackage{tabularx} 
\begin{document}

\title{GS2POSE: Marry Gaussian Splatting to 6D Object Pose Estimation}

\author{Junbo~Li,
        Weimin~Yuan,
        Yinuo~Wang,
        Yue~Zeng,
        Shihao~Shu,
        Cai~Meng,
        Xiangzhi~Bai
\thanks{ \textit{Co-first authors: Junbo Li, Weinin Yuan; Corresponding authors: Cai Meng,Tsai@buaa.edu.cn.}}
\thanks{Junbo Li, Weimin Yuan, Yinuo Wang, Yue Zeng, Shihao Shu, Cai Meng, Xiangzhi Bai are with the Image Processing Center, Beihang
University, Beijing 100191, China.}}

\maketitle

\begin{abstract}
Accurate 6D pose estimation of 3D objects is a fundamental task in computer vision, and current research typically predicts the 6D pose by establishing correspondences between 2D image features and 3D model features. However, these methods often face difficulties with textureless objects and varying illumination conditions. To overcome these limitations, we propose GS2POSE, a novel approach for 6D object pose estimation. GS2POSE formulates a pose regression algorithm inspired by the principles of Bundle Adjustment (BA). By leveraging Lie algebra, we extend the capabilities of 3DGS to develop a pose-differentiable rendering pipeline, which iteratively optimizes the pose by comparing the input image to the rendered image. Additionally, GS2POSE updates color parameters within the 3DGS model, enhancing its adaptability to changes in illumination. Compared to previous models, GS2POSE demonstrates accuracy improvements of 1.4\%, 2.8\% and 2.5\% on the T-LESS, LineMod-Occlusion and LineMod datasets, respectively. 
\end{abstract}


\IEEEpeerreviewmaketitle

\section{Introduction}

\IEEEPARstart{A}{ccurate} 6D object pose estimation is an essential task in the field of computer vision, with broad application in technologies such as robot navigation \cite{deng2020self,li2021latent}, image processing \cite{yuan2024mixed,yuan2025image,wang2025mamba,yuan2025guided,yuan2024simultaneous,wang2024samihs,li2024single,yuan2024weighted,yuan2023guided,wang2023coam,meng2023oscillating,li2022review,yuan2021efficient,yuan2019sgm} and virtual reality \cite{marchand2015pose,su2019deep}. Current mainstream pose estimation algorithms \cite{sun2022onepose,hai2023rigidity,he2022onepose++,chen2024zeropose,liu2022category} often rely on extracting 2D features from the input image and matching them with 3D features from the object model, which achieve high prediction accuracy by implementing attention mechanism that adeptly extract contextual information from images and seamlessly integrate it with high-quality object model.

However, acquiring high-precision object models of unseen objects can be quite challenge in real-world scenarios. To achieve object-model-free operation, some pose estimation algorithms \cite{luo2024object,vutukur2024nerf,li2023nerf,matteo20246dgs} leverage the breakthrough advancements in 3D reconstruction techniques, such as NeRF \cite{mildenhall2021nerf} or 3DGS \cite{kerbl20233d}. These reconstruction methods enable the real-time generation of 3D models using a series of pre-captured images, thus overcoming the dependence on object models. 

But these 3D-model-generation based pose estimation algorithms are highly sensitive to object textures, sensor type and environmental lighting. When the target object is textureless, these methods fail to establish accurate correspondences between image features and model features \cite{he2023contourpose}. In addition, these methods are almost implemented based on RGB images, which consequently leads to depth blur issues \cite{lin2023deep}. Furthermore, when there are changes in environmental lighting, it may lead to incorrect feature matching \cite{zhang2025robust}.

To enhance the robustness to textureless objects in 3D-model-generation based pose estimation algorithms, a novel pose estimation method named GS2POSE is proposed, which can also address the issues of depth blur and illumination adaptation.

To address the challenge of textureless objects, GS2POSE adopts a two-stage pose estimation approach: Coarse Pose Estimator and Pose Refiner. (1) In Coarse Pose Estimator stage, a novel coarse estimation network based on Normalized Object Coordinate Space (NOCS) representation is designed. It can provide a coarse pose estimation result of the new view, which possesses strong robustness to textureless objects. (2) In Pose Refiner stage, this paper introduces a method termed GS-Refiner, which leverages images to directly guide the refinement of object translation and rotation parameters. Different from the previous method, GS-Refiner relies on capturing the contour and color features of the objects to refine their poses, thereby reducing the dependency on texture features. Simultaneously, to address the issue of depth blur, GS2POSE designs a ray projection-based point cloud registration method, which aligns the point cloud generated by ray projection from 3DGS model with the point cloud produced by the RGBD image, effectively correcting depth prediction errors. Apart from that, GS2POSE updates the 3DGS model parameters related to color information, allowing it to adaptively adjust model brightness based on illumination change.

The contributions of the paper can be summarized as follows:

i) To enhance the robustness of our model in estimating the pose of textureless objects, we propose an innovative Pose Refinement Network. This network directly uses images to refine the object translation and rotation parameters, which relies on capturing the contour and color features of the objects to refine their poses, instead of the texture features. 

ii) To tackle the challenge of depth blur, we present a ray projection-based point cloud registration method. This approach aligns the point cloud generated by ray projection from 3DGS model with the point cloud produced by the RGBD image, effectively addressing the issue of depth prediction errors.

iii) To enhance adaptability to ambient lighting conditions, we develop an environment adoption strategy, which can update the 3DGS color parameters to adaptively adjust model brightness based on illumination change.

The reminder of this paper is organized as follows. Section 2 introduces the related work. Section 3 presents the proposed method. Section 4 provides the  experimental validation . Finally, Section 5 concludes this paper.

\section{Related Works}

In this section, we provide a brief summary of the development on 6D pose estimation, tracing the evolution from early approaches that utilized traditional methods to recent techniques that employ deep learning. Among the deep learning-based methods, we categorize them into two types based on whether they require feature from 3D models: 2D-2D feature-based and 2D-3D feature-based pose estimation methods. Within the realm of 2D-3D feature-based methods, a further distinction can be made between object-based and 3D-model-generation based methods.

\subsection{6D pose estimation based on traditional methods}

Traditional 6D pose estimation methods can be broadly categorized into feature matching-based, model-based, and geometry constraint-based approach. Feature matching-based methods employ SIFT or SURF algorithms to extract keypoints from images, then establish consistent feature correspondences across different viewpoints and utilize geometric constraints to refine pose estimation. Model-based methods, on the other hand, begin by constructing the object 3D model through Structure from Motion (SfM). Then the pose is iteratively optimized by minimizing the projection error between the target viewpoint and the 3D model. Geometry constraint-based methods optimize pose estimation by capturing the target object geometric properties then employing geometric constraints. Despite their effectiveness, these methods often face an obvious decline in accuracy when dealing with textureless objects or scenes with drastic lighting variations, which limits their practical applicability in real-world scenarios.

\subsection{6D pose estimation based on deep learning}

Due to the limited feature extraction and matching capabilities of traditional methods, convolutional neural networks (CNNs) have been increasingly used for pose estimation. Depending on whether features are extracted from 3D models, pose estimation methods can be divided into 2D-2D and 2D-3D feature-based approaches. Among 2D-3D feature-based methods, they can be further categorized into object-based and 3D-model-generation based methods.

\subsection{2D-2D feature-based pose estimation}
a convolutional neural network to predict the 2D or 3D bounding box of objects in an image. Subsequently, the pose is estimated from the bounding box corner points. For example, SSD-6D \cite{kehl2017ssd} predicts 2D bounding boxes and designs an evaluation network to select the best pose.  BB8 \cite{rad2017bb8} predicts the 3D bounding box based on the 2D image and recovers the 6D pose by back-projecting the eight corners. RADet \cite{li2020radet} adds a classifier to constrain the pose range, which can effectively resolve the problem of distinguishing similar viewpoints.

However, 2D-2D feature-based pose estimation methods heavily rely on the completeness of camera viewpoints in the training datasets. Moreover, these methods struggle to capture the overall structure of the target object, which limits their accuracy and robustness.

\subsubsection{2D-3D feature-based pose estimation}
To improve model accuracy and robustness, many methods use 3D object models to establish consistent correspondences between 2D image features and 3D model features. The PoseCNN model \cite{xiang2017posecnn} is designed with a dual-branch architecture that separately predicts the translational and rotational components of object pose, thereby facilitating a more effective decoupling of these pose parameters. The NOCS model \cite{wang2019normalized} transforms 3D models into a normalized object coordinate space, enabling the estimation of the 6D pose through the NOCS image. The onepose model \cite{sun2022onepose} first builds a sparse 3D object model via Structure-from-Motion (SfM) and incorporates a graph attention network into the feature matching module to improve feature extraction and correspondence estimation. The DeepIM model is designed with a neural network framework that iteratively optimizes pose estimation results by continuously comparing predicted images with ground truth images.

However, acquiring object models of unseen objects can be quite challenging in practice. With the advancement of 3D reconstruction techniques, methods based on NeRF \cite{mildenhall2021nerf} or 3DGS \cite{kerbl20233d} have been increasingly applied to the field of 6D pose estimation. These approaches leverage the strengths of model reconstruction bypass alleviating the dependency on object models. The Nerf-Pose model \cite{li2023nerf} employs a two-stage pose estimation strategy, introducing a NeRF-based PnP RANSAC method that effectively addresses the limitations of traditional PnP algorithms to achieve precise pose estimation. .

But these reconstruction-based methods exhibit poor robustness in predicting the textureless objects pose. To address above problem, we incorporate three positions and three rotation angles of the object as six learnable parameters into the 3DGS model. By computing reprojection errors, our proposed method achieves accurate pose estimation for textureless objects.

\section{The Proposed Method}\label{3}

\begin{figure*}[!th]
\centering
\includegraphics[width=1.0\linewidth]{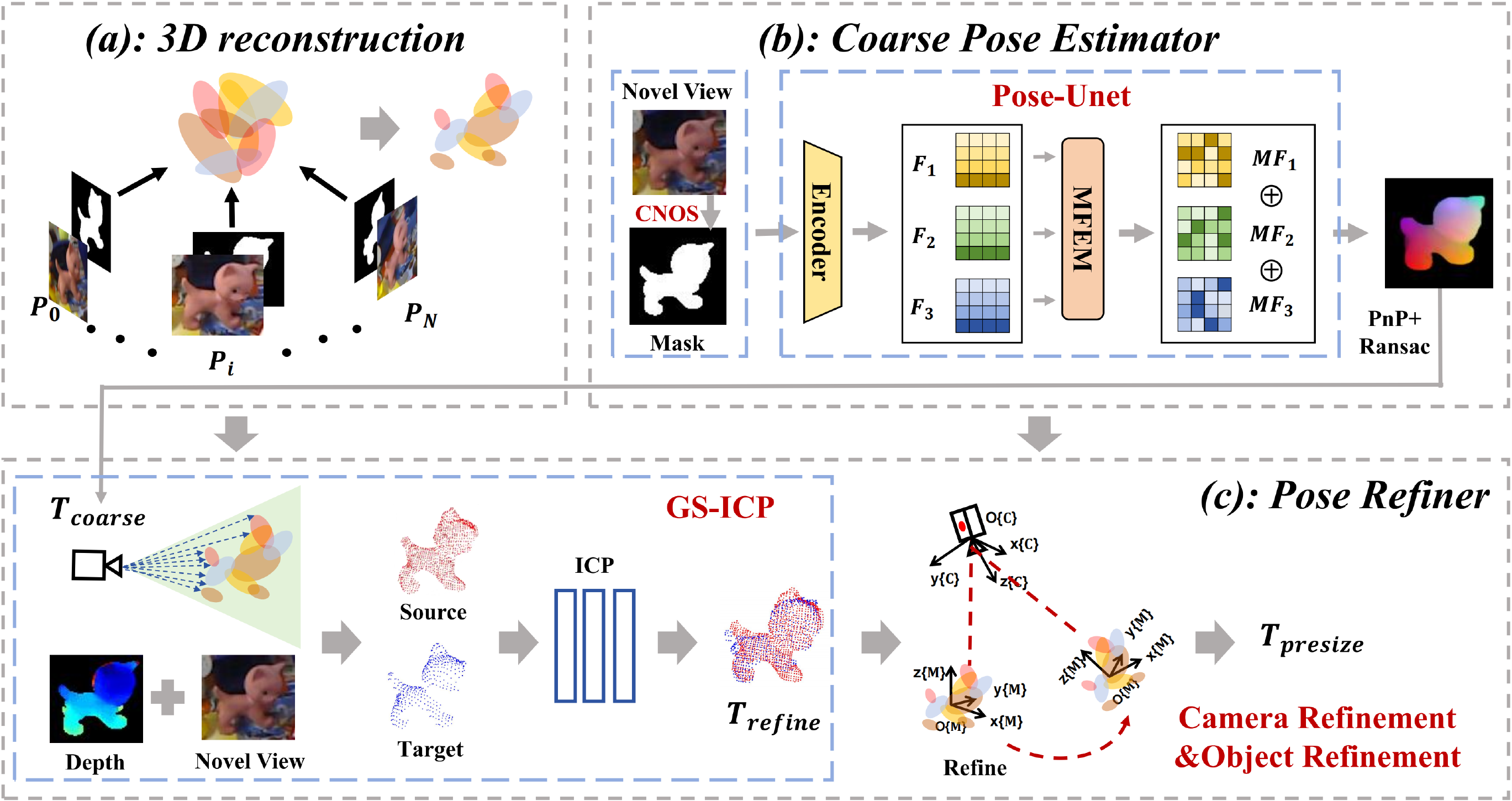}
\caption{The pipeline of the GS2POSE. \textit{(a) 3D reconstruction.} Before pose estimation, the 3D model of the object is constructed using 3DGS. \textit{(b) \underline{Coarse Pose Estimator}.} In this stage, the new view image and its corresponding mask (generated by CNOS model) are fed into the Pose-Unet. This network can generate the Normalized Object Coordinate Space (NOCS) image corresponding to the current viewpoint. By applying a PnP+RANSAC algorithm on the NOCS image, the coarse pose estimation transformation \( T_{coarse} \) can be obtained. \textit{(c) Pose Refiner.} In this stage, the point cloud generated from the RGBD image \(P_{target} \) and the point cloud generated by ray projection \(P_{source} \) are first processed by the ICP algorithm. ICP can perform an initial refinement of  \( T_{coarse} \), yielding  \( T_{refine} \). Subsequently,  \( T_{refine} \) is fed into the Camera Refinement and Object Refinement to get an accurate pose estimation result \( T_{precise} \).
}
\label{fig:model}
\end{figure*}

\subsection{Overview}

The overall architecture of GS2POSE model is illustrated in Fig.\ref{fig:model}. Prior to pose estimation, the target object is captured from multiple viewpoints, followed by 3D reconstruction using 3DGS algorithm. During pose estimation, GS2POSE utilizes the
new perspective 
RGB image  \( I_{\mathrm{rgb}} \), along with its corresponding mask  \( I_{\mathrm{mask}} \) (generated by CNOS model) and depth map \( I_{\mathrm{depth}} \) as inputs. These inputs are initially processed by the Coarse Pose Estimator to obtain a coarse pose estimation transformation \( T_{\mathrm{coarse}} \). Subsequently, \( T_{\mathrm{coarse}} \) is fed into the Pose Refiner for iterative refinement, yielding an accurate pose prediction transformation \( T_{\mathrm{precise}} \).

\subsection{3D Gaussian Splatting}

3D Gaussian Spheres (3DGS) is a scene representation method that describes objects in the world coordinate system using Gaussian spheres. All parameters of the 3D Gaussian Spheres are learnable, including the position parameters \( \mu \), opacity \( a \), the 3D covariance matrix \( r \), and the spherical harmonics \( sh \). Given any point \( x \) in the world coordinate system, the 3D Gaussian sphere defined at point \( x \) according to the Gaussian distribution is as follows:
\begin{align} 
\begin{split}
f(x; \mu, \Sigma) = \exp \left( -\frac{1}{2} (x - \mu)^{\mathrm{T}} \Sigma^{-1} (x - \mu) \right)
\end{split} 
\end{align}

\begin{align} 
\begin{split}
\Sigma = R S S^{\mathrm{T}} R^{\mathrm{T}}
\end{split} 
\end{align}

where \( R \) denotes the rotation matrix computed from \( r \), and \( S \) represents the diagonal matrix derived from \( s \). Subsequently, a fast rasterization approach is employed to project the 3D Gaussian points onto a 2D plane for rendering.

\subsection{Coarse Pose Estimator}

To enhance the generalization capability for textureless objects while providing the Pose Refiner with a relatively accurate initial pose prediction result, a lightweight NOCS image generation network ( Pose-Unet ) has been designed to predict the coarse pose. Pose-Unet adopts the ResNet50 as encoder. During the down sampling process, three distinct feature layers of varying scales are extracted, labeled \( F_{1} \) (128 $\times$ 128), \( F_{2} \) (64 $\times$ 64) and \( F_{3} \) (32 $\times$ 32). Considering the target objects is textureless, Pose-Unet proposes a multi-scale feature enhancement module (MFEM). The MFEM comprises three dilated convolution layers, which have kernel sizes of $3 \times3$, $3 \times3$, and $5 \times5$, with corresponding dilation rates of 2, 4, and 2, respectively. After feature fusion, the Polarized self-attention (PSA) (Liu et al. 2021) is incorporated to further enhance the capture of contextual features. 

Then the three feature layers  \( MF_{1} \), \( MF_{2} \), and \( MF_{3} \) obtained from the MFEM module are upsampled to the original image size using deconvolution and subsequently merged. Finally, Pose-Unet can generate the NOCS image corresponding to the new view image. The loss function used in Pose-Unet is designed as follows: 
\begin{align}
\begin{split}
Loss_{coarse} &= L_{1}| I_{nocs}*I_{mask}- I_{gen}*I_{mask}|.
\end{split}
\end{align}

In the aforementioned equation, \( I_{nocs} \) represents the real NOCS image rendered by the 3DGS model from the current perspective, while \( I_{gen} \) denotes the NOCS image generated by Pose-Unet. Subsequently, pixels with brightness exceeding a predetermined threshold are selected from the NOCS image, which are then fed into the PnPRansac function. Finally, the coarse pose estimation  \( T_{coarse} \) can be abtained.

\subsection{Pose Refiner}

Due to the limited accuracy of \( T_{coarse} \) obtained from Pose-Unet, a multi-stage refinement algorithm termed GS-refiner is designed, which leverages the 3DGS to represent object. Meanwhile, this algorithm employs an iterative reprojection method to provide a precise pose estimation. The GS-refiner is specifically divided into three stages: GS-ICP; Camera Refiner and Object Refiner; GS-light.

\subsubsection{\textbf{GS-ICP}}

To rectify the prediction bias in the depth direction of \( T_{coarse} \), the GS-ICP algorithm is proposed. The algorithm initially generates a target point cloud model \(P_{target} \) based on the new perspective RGBD image. Subsequently, it observes the 3DGS point cloud model from the \( T_{coarse} \). GS-ICP denotes the point clouds generated by the 3DGS as \(P_{model} \), represented as a set of points $P = \{ p_1, p_2, \ldots, p_n \}$ in the 3D space, where each point $p_k$ is defined by its coordinates $(x_k, y_k, z_k)$. For each pixel in the RGB image, a ray \( R \) is cast from the camera optical center \( C \) in the direction of the pixel corresponding 3D coordinate. The ray can be defined parametrically as:
\begin{equation}
R(t) = C + t \cdot D .
\end{equation}
Where \( D \) is the direction vector of the ray and \( t \) is a scalar parameter representing the distance along the ray. For each ray, to find the points \( p_{int} \) that are intersected by the ray within a certain threshold distance \( \epsilon \), The formula can be expressed as:

\begin{equation}
\text{Distance}(R(t), p_k) = \min_{t \geq 0} \| R(t) - p_k \| < \epsilon.
\end{equation}

The closest point \( p_{\text{closest}} \) is then selected based on the minimum distance:
\begin{equation}
p_{\text{closest}} = \arg\min_{p_k \in P_{int}} \|  C - p_k \|.
\end{equation}

After identifying the nearest points along each ray, the source point cloud \(P_{source} \) can be obtained. Subsequently, by performing point cloud registration ICP between \(P_{source} \) and \(P_{target} \),  depth information can be refined, 

\subsubsection{\textbf{Camera refiner and Object refiner}}

\begin{figure}[!t]
\centering

\includegraphics[width=1.0\linewidth]{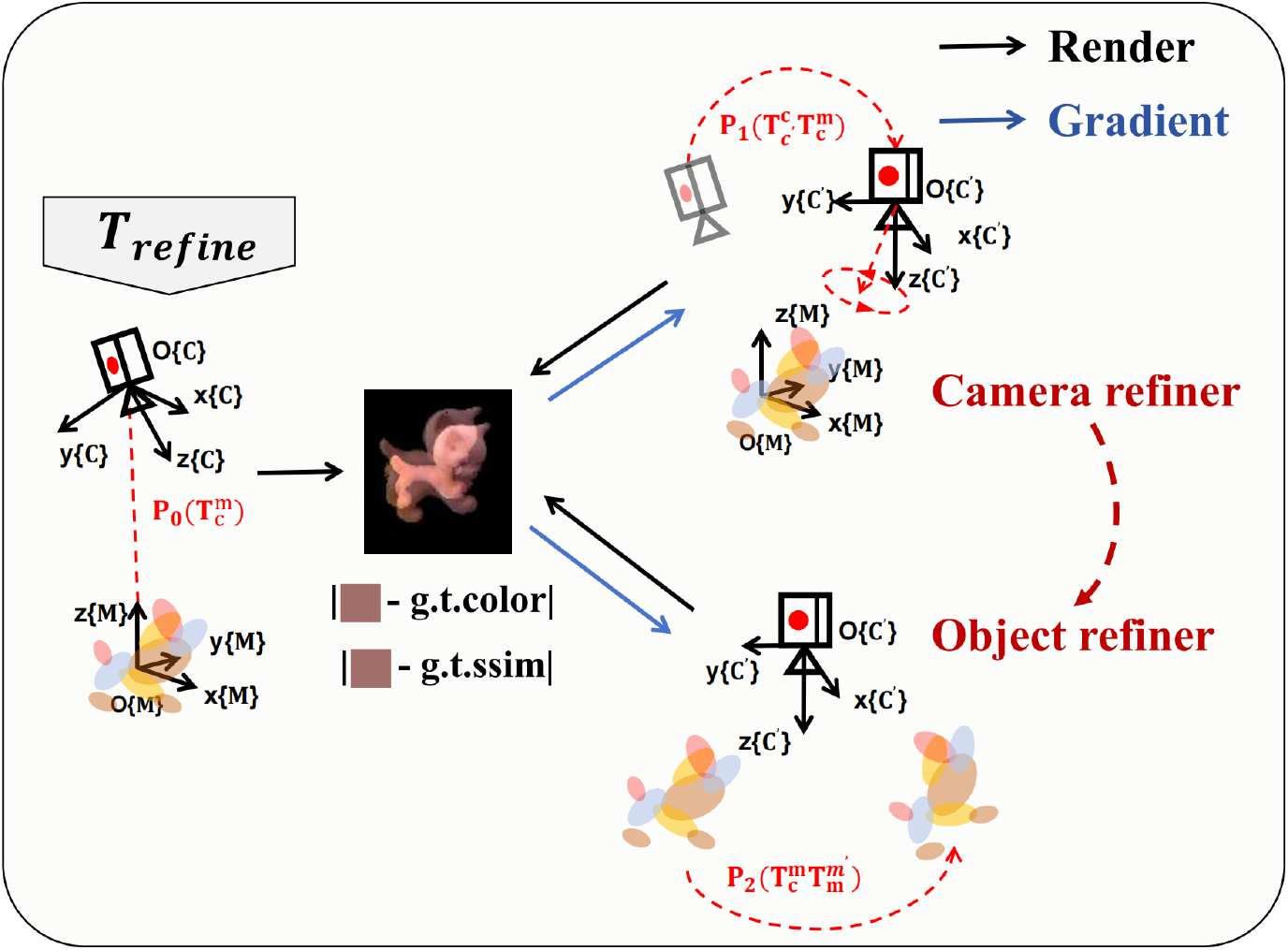}

\caption{The principle of the Camera refiner and Object refiner. Upon inputting the pose estimation transformation \(T_{refine} \), the translation parameters undergo initial refinement through the Camera Refiner. Subsequently, the rotation parameters are corrected during the Object Refiner phase.}
\label{fig:refine network}
\end{figure}

Following the correction of depth information, GS-refiner subsequently refined the translation and rotation parameters. Inspired by 3D Gaussian Splatting SLAM \cite{yan2024gs}, Lie algebra is used to represent pose changes between coordinate systems. Since this method is designed for textureless objects, a strategy based on reprojection is proposed to correct translation and rotation errors.  


In contrast to methods that solely rely on moving the object to refine the pose, GS-refiner consists of two steps: the Camera Refiner and the Object Refiner. The Camera Refiner keeps the object fixed while adjusting the camera position to refine the translation parameters. Subsequently, the Object Refiner keeps the camera in place while rotating the object to correct the rotation parameters. The principle of the Camera refiner and Object refiner is illustrated in Fig. \ref{fig:refine network}.

\textbf{A. Design Principles}

The reason for utilizing both the Camera Refiner and the Object Refiner is that, when optimizing translation parameters, rotating the camera offers two key advantages over moving the object.

(1) Due to the inherent iterative optimization properties of the Lie algebra, the optimization process of the translational parameters \(x\) and \(y\) both alters the depth information \(z\). Consequently, when the object is moved to correct the translational parameters, the distance between the object's centroid and the camera's optical center is modified, leading to changes in the depth information. In contrast, when the camera is rotated to adjust the translational parameters, the distance between the object's centroid and the camera's optical center remains constant. It can reduce the errors arising from variations in depth information.

(2) Due to the fact that the distance between the camera and the object is often significantly greater than the object diameter, a slight rotation of the camera pose can result in a huge translation of the object, thereby accelerating the iterative optimization process.

After correcting the translation parameters, if the rotation parameters are subsequently adjusted by rotating the camera, significant changes in the translation parameters may occur. In contrast, modifying the rotation parameters by rotating the object can preserve the accuracy of the translation parameters. 

\textbf{B. Mathematical Principles}

GS-Refiner objective is to compute the transformation matrix \(\tilde{{T}^{m}_{c}}\) from the object coordinate system \{m\} to the camera coordinate system \{c\}, which includes a translation vector \(t^{m}_{c}\) and a rotation matrix \(R^{m}_{c}\), as follows:
\begin{align}
\begin{split}
\tilde{T}^{m}_{c} &= \left[ \begin{array}{cc}
R^{m}_{c} & t^{m}_{c} \\
0^{T} & 1
\end{array} \right], \quad 
t^{m}_{c} = \begin{pmatrix} x^{m}_{c} & y^{m}_{c} & z^{m}_{c} \end{pmatrix}^{T}.
\end{split}
\end{align}

For continuous 3D representation, the mean \( p_m \) and covariance $\Sigma_{m}$ defined in the object coordinate system \{m\}, represent the Gaussian position and its ellipsoidal shape. The 3D Gaussian (\( p_m \) $\Sigma_{m}$) in object coordinate are related to the 2D Gaussian (\( p_i \) $\Sigma_{i}$) on the image plane through a projective transformation:

\begin{align} 
\begin{split}
p_i = \pi\begin{pmatrix} {T}^{m}_{c}\cdot p_m \end{pmatrix}, 
\Sigma_i = J {R}^{m}_{c} \Sigma_{m} \left( {R}^{m}_{c} \right)^T J^{T},
\end{split} 
\end{align}

where $\pi$ is the projection operation and J is the Jacobian of the linear approximation of the projective transformation. The coordinates of any object point in the camera coordinate system are represented as \( p_c \). According to the chain rule of differentiation:

\begin{align}
    \frac{\partial p_i}{\partial \tilde{T}^{m}_{c}} &= \frac{\partial p_i}{\partial p_c} \frac{\partial p_c}{\partial \tilde{T}^{m}_{c}}, \\
    \frac{\partial \Sigma_i}{\partial \tilde{T}^{m}_{c}} &= \frac{\partial \Sigma_i}{\partial J} \frac{\partial J}{\partial p_c} \frac{\partial p_c}{\partial \tilde{T}^{m}_{c}} + \frac{\partial \Sigma_i}{\partial{R}^{m}_{c}} \frac{\partial {R}^{m}_{c}}{\partial \tilde{T}^{m}_{c}}, 
\end{align}

the expressions \(\frac{\partial p_c}{\partial \tilde{T}^{m}_{c}}\) and \(\frac{\partial {R}^{m}_{c}}{\partial \tilde{T}^{m}_{c}}\) cannot be directly obtained through differentiation, which require further derivation.

In the Camera Refiner, the object remains stationary while the camera undergoes rotation. During each iteration, a left perturbation \( \tilde{T}^{c}_{c'} \in SE(3) \) will be applied to \( \tilde{T}^{m}_{c} \). The object points coordinates in the camera coordinate system are represented as \(p_{c'}\) after perturbation.
\begin{align}
p_{c'} = \tilde{T}^{c}_{c'} \tilde{T}^{m}_{c} p_m = \tilde{{T}^{c}_{c'}} p_c.
\end{align}

Let the Lie algebra corresponding to \( \tilde{T}^{c}_{c'} \) be denoted as:
\begin{align} 
\begin{split}
\tau_c = \begin{bmatrix} \rho_c & \phi_c \end{bmatrix}^T \in se(3), \quad p_{c'} = \exp(\tau_c) p_c.
\end{split} 
\end{align}

At this point, let \( p_c \) be differentiated with respect to \( \tau_c \):
\begin{align}
\frac{\partial p_c}{\partial \tau_c} 
&= \lim_{\tau_c \to 0} \frac{\text{Exp}(\tau_c)p_c - p_c}{\tau_c} \\
&= \lim_{\tau_c \to 0} \frac{(I + \tau_c\hat{})p_c - p_c}{\tau_c} \\
&= \lim_{\tau_c \to 0} \frac{\tau_c\hat{}p_c}{\tau_c} \\
&= \lim_{\tau_c \to 0} \frac{\begin{bmatrix}
\phi_c\hat{} & \rho_c \\
0^T & 0
\end{bmatrix}p_c}{\tau_c} \\
&= \lim_{\tau_c \to 0} \frac{\phi_c\hat{} p_c^{:3} + \rho_c}{\tau_c} \\
&= \lim_{\tau_c \to 0} \frac{-p_c^{:3\hat{}}\phi_c + \rho_c}{\tau_c} \\
&= [I, -p_c^{:3\hat{}}].
\end{align}

On the other hand, let \(R^{m}_{c}\) be differentiated with respect to \( \phi_c \):

\begin{align}
\frac{\partial{R}^{m}_{c}}{\partial \phi_c} 
&= \lim_{\phi_c \to 0} \frac{\text{Exp}(\phi_c){R}^{m}_{c} -{R}^{m}_{c}}{\phi_c} \\
&= \lim_{\phi_c \to 0} \frac{(I + \phi_c\hat{}) {R}^{m}_{c} - {R}^{m}_{c}}{\phi_c} \\
&= \lim_{\phi_c \to 0} \frac{\phi_c\hat{} {R}^{m}_{c}}{\phi_c} \\
&= [ {-R_{c}^{m}}_{:,1}\hat{}, {-R_{c}^{m}}_{:,2}\hat{}, {-R_{c}^{m}}_{:,3}\hat{}]^T
\end{align}

Subsequently, we can obtain the calculation results for \( \tilde{T}^{c}_{c'} \).

In the Object Refiner, the camera remains stationary while the object undergoes rotation. During each iteration, a right perturbation \( \tilde{T}^{m'}_{m} \in SE(3) \) will be applied to \( \tilde{T}^{m}_{c} \). Since each object point \( p_m \) on the object is rigidly attached to the object coordinate system, its coordinate values in the object coordinate system will not change, that is:
\begin{align} 
\begin{split}
p_c = \tilde{T}^{m}_{c} \tilde{T}^{m'}_{m} p_m.
\end{split} 
\end{align}

Let the Lie algebra corresponding to \( \tilde{T}^{m'}_{m} \) be denoted as:
\begin{align} 
\begin{split}
\tau_m = \begin{bmatrix} \rho_m & \phi_m \end{bmatrix}^T \in se(3), \quad p_c = \tilde{T}^{m}_{c} \exp(\tau_m) p_m.
\end{split} 
\end{align}

At this point, let \( p_c \) be differentiated with respect to \( \tau_m \):

\begin{align}
\frac{\partial p_c}{\partial \tau_m} 
&= \lim_{\tau_m\to 0} \frac{\tilde{T}^{m}_{c}\text{Exp}(\tau_m)p_m - \tilde{T}^{m}_{c}p_m}{\tau_m} \\
&= \lim_{\tau_m \to 0} \frac{\tilde{T}^{m}_{c}(I + \tau_m\hat{})p_m - \tilde{T}^{m}_{c}p_m}{\tau_m} \\
&= \lim_{\tau_m \to 0} \frac{\tilde{T}^{m}_{c}\tau_m\hat{}p_m}{\tau_m} \\
&= \lim_{\tau_m \to 0} \frac{\tilde{T}^{m}_{c}\begin{bmatrix}
\phi_m\hat{} & \rho_m \\
0^T & 0
\end{bmatrix}p_m}{\tau_m} \\
&= \lim_{\tau_m \to 0} \frac{\tilde{T}^{m}_{c}\phi_m\hat{} p_m^{:3} + \tilde{T}^{m}_{c}\rho_m}{\tau_m} \\
&= [\tilde{T}^{m}_{c} \cdot I, -\tilde{T}^{m}_{c}p_c^{:3\hat{}}].
\end{align}

On the other hand, let \(R^{m}_{c}\) be differentiated with respect to \( \phi_c \):

\begin{align}
\frac{\partial{R}^{m}_{c}}{\partial \phi_m} 
&= \lim_{\phi_m \to 0} \frac{{R}^{m}_{c}\text{Exp}(\phi_m) -{R}^{m}_{c}}{\phi_m} \\
&= \lim_{\phi_m \to 0} \frac{{R}^{m}_{c} (I + \phi_m\hat{})- {R}^{m}_{c}}{\phi_m} \\
&= \lim_{\phi_m \to 0} \frac{ {R}^{m}_{c}\phi_m\hat{}}{\phi_m} \\
&= [ {-R_{c}^{m}}_{1,:}\hat{}, {-R_{c}^{m}}_{2,:}\hat{}, {-R_{c}^{m}}_{3,:}\hat{}]^T
\end{align}

Subsequently, we can obtain the calculation results for \( \tilde{T}^{m}_{m'} \).

\subsubsection{\textbf{GS-light}} 

To address issues such as reflections and shadows under varying lighting conditions, GS-refiner incorporated environment adoption during Object refiner. In the 3DGS model, colors are represented using spherical harmonics. The mathematical principle of spherical harmonics is expressed as follows:
\begin{equation}
    f(t) = \sum_{l} \sum_{m = -l}^{l} c_{l}^{m} y_{l}^{m}(\theta, \phi).
\end{equation}

In the above formula, \( (\theta, \phi) \) represents the directional information of the viewing angle. \( y_{l}^{m} \) represents the basis functions,  \( c_{l}^{m} \) represents the coefficient. Following the 3DGS model, GS-light sets \( l \) to 3. Then GS-light get 16 spherical harmonic parameters of the Gaussian model as learnable parameters. Additionally, GS-light lock other parameters, such as the scale parameter of the Gaussian spheres, their position parameters \(xyz\) relative to the object coordinate system, and their transparency. This is to prevent the model from compromising its original structure during iterations. Such compromises could negatively impact the accuracy of angle estimation.

\begin{table}[!thbp]
\centering
\small
\begin{tabular}{c|c|c|c}
\toprule
Depth & Method & Publish & VSD(\%) \\
\midrule
& Pix2Pose & [ICCV2019] & 29.5 \\
& PVNet & [CVPR2019] & 40.4 \\
& CosyPose & [ECCV2020] & 63.8 \\
\midrule
\multirow{5}{*}{$\bullet$} & DenseFusion & [CVPR2019] & 10.0 \\
& MP-Encoder & [IJCV2020] & 69.5 \\
& OVE6D & [CVPR2022] & 69.4 \\
& OVE6D+SDF & [ICLR2023] & \underline{73.3} \\
& SS-Pose & [TII2024] & 67.8 \\
\midrule
$\bullet$ & GS2POSE & & \textbf{74.7(1.4$\uparrow$)}\\
\bottomrule
\end{tabular}

\caption{Experiments on the T-LESS Dataset. Underlined and bold values represent the second-best and best results, respectively.}
\label{tab:T-LESS}
\end{table}

\begin{table*}[!thbp] 
\centering
\setlength{\tabcolsep}{1mm} 
\small 
\begin{tabular}{ c|c|c| *{8}{c}|c}
\toprule
Depth
& Mehod
& Publish
& $\text{ape}$ 
& $\text{can}$ 
& $\text{cat}$ 
& $\text{driller}$ 
& $\text{duck}$ 
& $\text{eggbox*}$ 
& $\text{glue*}$ 
& $\text{holep}$
& $\text{avg}$
\\
\midrule
\multirow{5}{*}{} 
& PoseCNN & [RSS2018] & 9.6 & 45.2 & 0.9 & 41.4 & 19.6 & 22.0 & 38.5 & 22.1 & 24.9 \\
& Pix2Pose & [ICCV2019] & 22.0 & 44.7 & 22.7 & 44.7 & 15.0 & 25.2 & 32.4 & 49.5 & 32.0 \\
& PVNet & [CVPR2019] & 15.8 & 63.3 & 16.7 & 65.7 & 25.2 & 52.0 & 51.4 & 45.6 & 40.8 \\
& HybridPose & [CVPR2020] & 20.9 & 75.3 & 24.9 & 70.2 & 27.9 & 52.4 & 53.8 & 54.2 & 47.5 \\
& NVR-Net & [TCSVT2023] & 43.1 & 82.9 & 27.2 & 69.7 & 44.2 & 49.7 & \underline{74.3} & 61.7 & 56.6 \\
\midrule
\multirow{7}{*}{$\bullet$} 
& PVN3D  & [CVPR2020] & 33.9 & 88.6 & 39.1 & 78.4 & 41.9 & \textbf{80.9} & 68.1 & 74.7 & 63.2 \\
& PR-GCN & [ICCV2021] & 40.2 & 76.2 & \underline{57.0} & 82.3 & 30.0 & 68.2 & 67.0 & \textbf{97.2} & 65.0 \\
& FFB6D & [CVPR2021] & 47.2 & 85.2 & 45.7 & 81.4 & 53.9 & 70.2 & 60.1 & 85.9 & 66.2 \\
& BRD6D & [TASE2023] &  - & - & - & - & - & - & - & - & 66.8 \\
& DeepFusion & [TASE2024]  &  56.6 & \textbf{95.8} & 40.0 & \textbf{93.5} & \underline{57.1} & 70.0 & 56.7 & \underline{87.0} & \underline{69.6} \\
& PCKRF & [TVCG2025] &\underline{63.8} & \underline{94.8} & 39.8 & \underline{82.3} & 50.5 & 63.7 & 70.8 & 83.7 & 68.7 \\
\midrule
\multirow{1}{*}{$\bullet$} 
 & GS2POSE  & & \textbf{70.8(7.0$\uparrow$)}  & 83.9 & \textbf{57.4(0.4$\uparrow$)} & 78.0 & \textbf{60.0(2.9$\uparrow$)} & \underline{75.8} & \textbf{74.8(0.5$\uparrow$)} & 79.5  & \textbf{72.4(2.8$\uparrow$)} \\
\bottomrule
\end{tabular}

\caption{
Experiments on the LineMOD-Occlusion Dataset. The results marked with * indicate symmetric objects. Underlined and bold values represent the second-best and best results, respectively.
}
\label{tab:LineMOD-Occlusion}
\end{table*}

\subsubsection{\textbf{Loss Fuction}} 

Referring to the design of the original 3DGS loss function, the loss function in this method is designed as follows:
\begin{small}
\begin{align}
\begin{split}
\text{loss}(I_{\mathrm{in}}, I_{\mathrm{pred}}) &= \lambda L_{image} + (1 - \lambda) L_{\mathrm{dssim}}+\beta* L_{depth}.
\end{split}
\end{align}
\end{small}

In the aforementioned equation, \( I_{\mathrm{in}} \) represents the input rgb image and depth image, while \( I_{\mathrm{pred}} \) denotes the color image and depth image obtained through 3DGS rendering. \( L_{image} \) signifies the \( L_{1} \) loss between the ground truth color image and the predicted color image. \( L_{\mathrm{dssim}} \) indicates the structural similarity loss between the ground truth color image and the predicted color image. \( L_{depth} \) represents the \( L_{1} \)  loss between the ground truth depth image and the predicted depth image. 

\begin{figure*}[!t]
\centering
\includegraphics[width=1.0\linewidth]{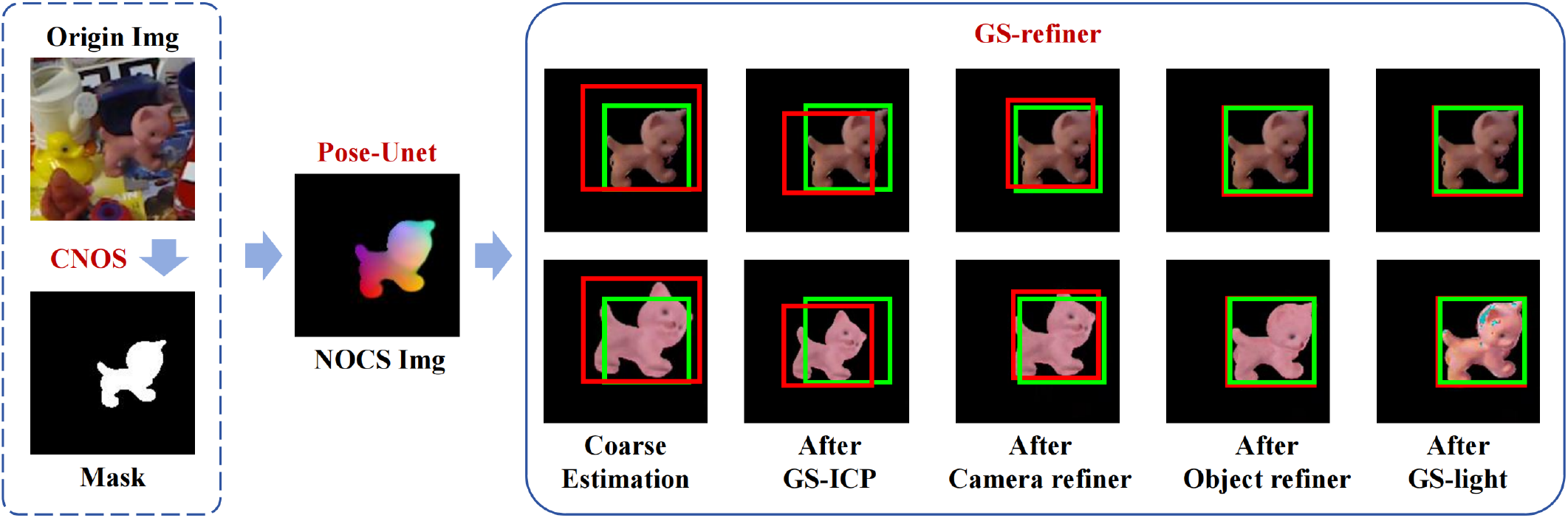}
\caption{Visualization of the GS2POSE process. The original image is first processed through the CNOS model to generate the corresponding mask image. Subsequently, both the original image and the mask are input into the model, which are then processed by the coarse Pose-Unet to produce the NOCS image. The coarse pose estimation transformation can be directly derived from the NOCS images. Following this, the coarse pose estimation transformation is utilized as input for the GS-refiner to achieve precise pose estimation. In the GS-refiner stage, the images presented in the top row represent real-view images, while those in the bottom row are rendered outputs generated by the GS-refiner.
}
\label{fig:refine_liucheng}
\end{figure*}

\section{Experimental Results and Analyses}

\begin{figure}[!t]
\centering
\resizebox{0.7\linewidth}{!}{
\includegraphics[width=\linewidth]{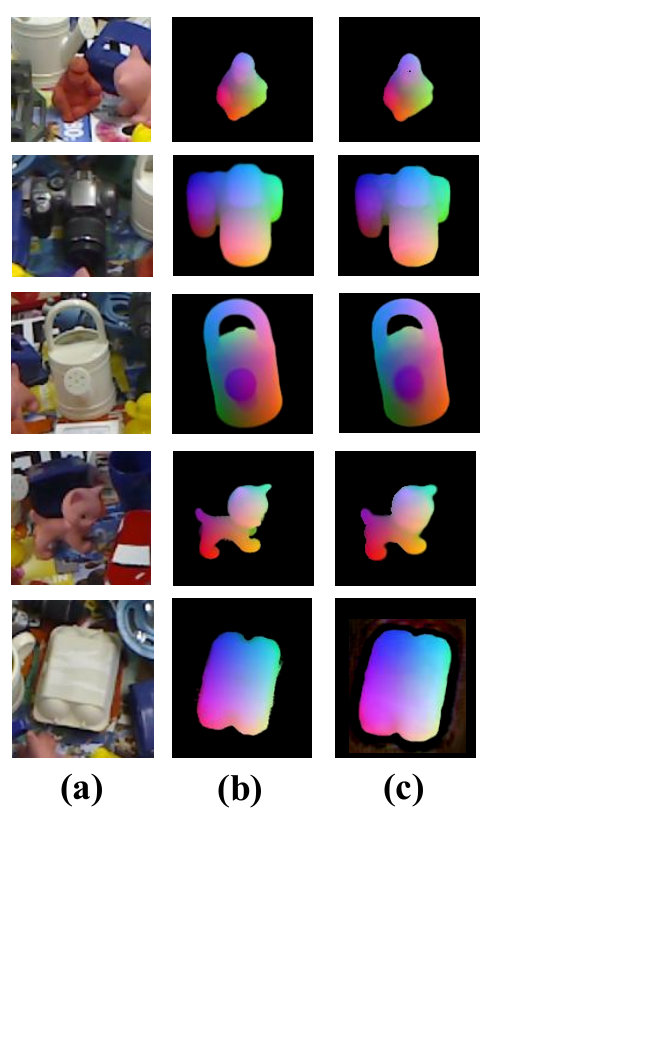}
}
\caption{
Visualization Results of the Pose-Unet Network: (a) represents the original images, (b) represents the ground truth of the NOCS images, (c) represents the NOCS images generated by the Pose-Unet network.
}
\label{fig:coarse_result}
\end{figure}

\begin{figure}[!t]
\centering

\includegraphics[width=0.9\linewidth]{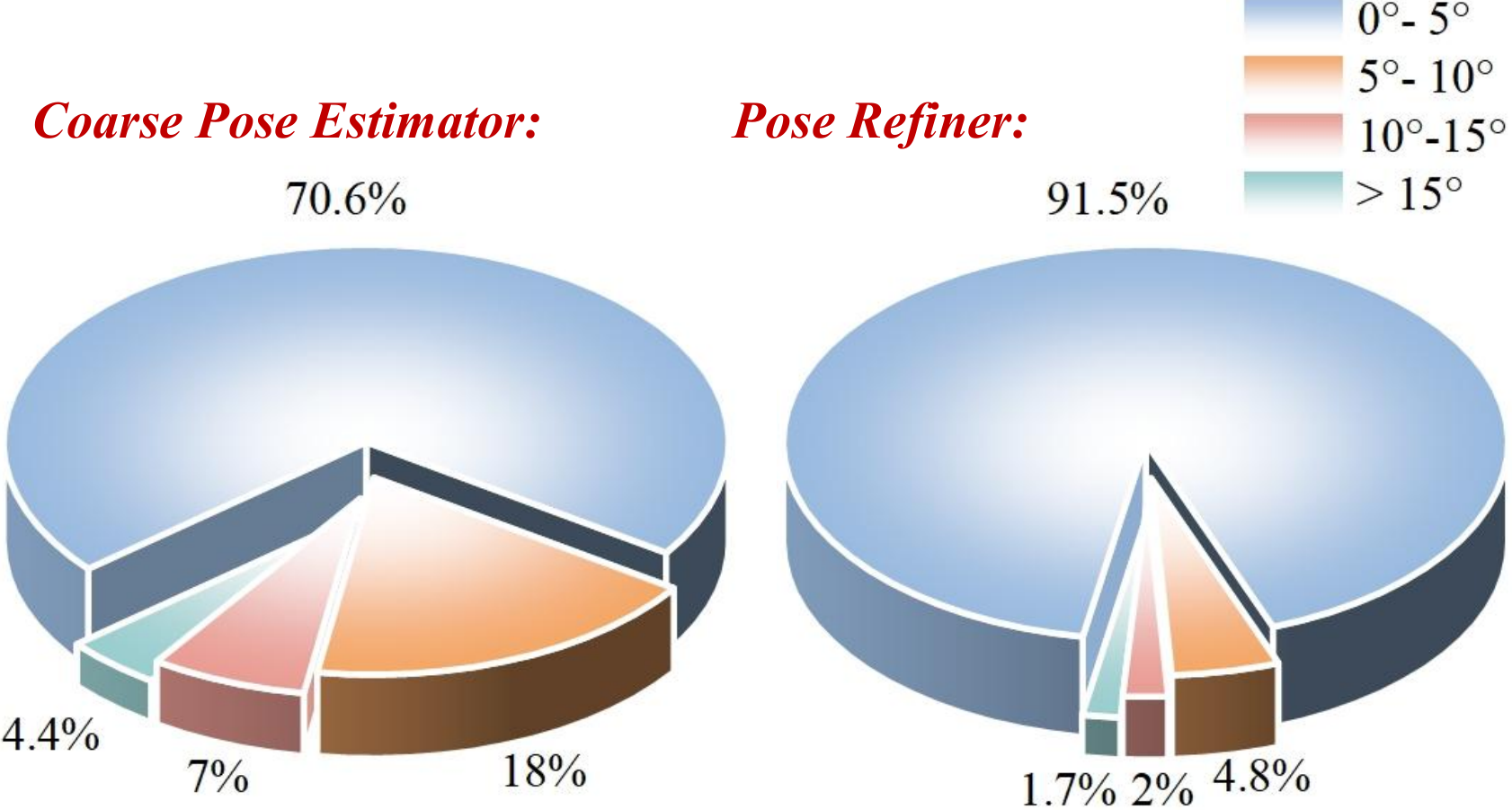}

\caption{
The angular prediction error distributions of the Coarse Pose Estimator and the Pose Refiner.
}
\label{fig:fenbu}
\end{figure}

\begin{table*}[!thbp] 
\centering
\setlength{\tabcolsep}{1mm} 
\small 
\begin{tabular}{c|c|c| *{13}{c}|c}
\toprule
Depth
& Method
& Publish
& $\text{ape}$ 
& $\text{bvise}$ 
& $\text{cam}$ 
& $\text{can}$ 
& $\text{cat}$ 
& $\text{driller}$ 
& $\text{duck}$ 
& $\text{eggbox*}$ 
& $\text{glue*}$ 
& $\text{holep}$
& $\text{iron}$
& $\text{lamp}$
& $\text{phone}$
& $\text{avg}$
\\
\midrule
\multirow{6}{*}{} 
& PoseCNN  & [RSS2018] & 77 & 97.5 & 93.5 & 96.5 & 82.1 & 95 & 77.7 & 97.1 & 99.4 & 52.8 & 98.3 & 97.5 & 87.7 & 88.6
\\
& PVNet  & [CVPR2019] &  43.6 & 99.9 & 86.9 & 95.5 & 79.3 & 96.4 & 52.6 & 99.2 & 95.7 & 81.9 & 98.9 & 99.3 & 92.4 & 86.3
\\
& CDPN & [ICCV2019]  &  64.4 & 97.8 & 91.7 & 95.9 & 83.8 & 96.2 & 66.8 & 99.7 & 99.6 & 85.8 & 97.9 & 97.9 & 90.8 & 89.9
\\
& DPOD  & [ICCV2019] &  87.7 & 98.5 & 96.1 & 99.7 & 94.7 & 98.8 & 86.3 & 99.9 & 96.8 & 86.9 & 100 & 96.8 & 94.7 & 95.1
\\
& NVRNet & [TCSVT2023] &  83.3 & 98.8 & 94.9 & 98.2 & 95.4 & 98.3 & 85.2 & 99.9 & 99.6 & 91.5 & 98.6 & 99.6 & 94.8 & 95.3
\\
& NeRF-Pose & [ICCV2023]  & 89.1 & 99.3 & 98.7 & 99.1 & 97.1 & 97.4 & 90.3 & 99.6 & 98.1 & 94.3 & 98.1 & 97.9 & 96.4 & 96.6
\\
\midrule
\multirow{11}{*}{$\bullet$} 
& PointFusion  & [CVPR2018] & 70.4 & 80.7 & 60.8 & 61.1 & 79.1 & 47.3 & 63 & 99.9 & 99.3 & 71.8 & 83.2 & 62.3 & 78.8 & 73.7
\\
& DenseFusion &[CVPR2019]  &  92.3 & 93.2 & 94.4 & 93.1 & 96.5 & 87.0 & 92.3 & 99.8 & 100 & 92.1 & 97 & 95.3 & 92.8 & 94.3
\\
& G2L-Net & [CVPR2020] &  96.8 & 96.1 & 98.2 & 98.0 & 99.2 & 99.8 & 97.7 & \textbf{100.0} & \textbf{100.0} & 99.0 & 99.3 & 99.5 & 98.9 & 98.7
\\
& PVN3D & [CVPR2020] & 97.3 & 99.7 & 99.6 & 99.5 & \underline{99.8} & 99.3 & 98.2 & 99.8 & \textbf{100.0} & \textbf{99.9} & 99.7 & \underline{99.8} & 99.5 & 99.4
\\
& FFB6D & [CVPR2021] & \underline{98.4} & \textbf{100.0} & \underline{99.9} & \textbf{99.8} & \textbf{99.9} & \textbf{100.0} & 98.4 & \textbf{100.0} & \textbf{100.0} & 99.8 & \textbf{99.9} & \textbf{99.9} & \underline{99.7} & \underline{99.7}
\\
& FS6D & [CVPR2022] &  74.0 & 86.0 & 88.5 & 86.0 & 98.5 & 81.0 & 68.5 & \textbf{100.0} & 99.5 & 97.0 & 92.5 & 85.0 & 99.0 & 88.9
\\
& FS6D+ICP & [CVPR2022] &  78.0 & 88.5 & 91.0 & 89.5 & 97.5 & 92.0 & 75.5 & 99.5 & 99.5 & 96.0 & 87.5 & 97.0 & 97.5 & 91.5
\\
& BDR6D & [TASE2023] & - & - & - & - & - & - & - & - & - & - & - & - & - & 99.5
\\
& HFT6D & [TASE2024] & 97.4 & 98.9 & 98.8 & 98.7 & 98.9 & 99.8 & \underline{98.7} & \textbf{100.0} & \textbf{100.0} & 98.9 & 99.3 & 99.5 & 98.9 & 99.1
\\
\midrule
$\bullet$ & GS2POSE &  & \textbf{98.8} & \textbf{100.0} & \textbf{100.0} & \textbf{99.8} & \underline{99.8} & \underline{99.9} & \textbf{98.9} & \textbf{100.0} & \textbf{100.0} & \textbf{99.9} & \textbf{99.9} & \underline{99.8} & \textbf{99.8} & \textbf{99.8}
\\
\bottomrule
\end{tabular}

\caption{Experiments on the LineMOD Dataset. The results marked with * indicate symmetric objects. Underlined and bold values represent the second-best and best results, respectively.}
\label{tab:LineMOD}
\end{table*}

\begin{table*}[!thbp]
\centering
\caption{
Effects of different components in GS2POSE. Experiments are conducted on selected objects from the LM dataset.
}
\label{tab:ablation}
\setlength{\tabcolsep}{1pt} 
\small
\begin{tabular}{c|cccc|*{13}{c}|c}
\toprule
\multirow{2}{*}{Coarse Est.} & \multicolumn{4}{c|}{Refine Est.} & $\text{ape}$ & $\text{bvise}$ & $\text{cam}$ & $\text{can}$ & $\text{cat}$ & $\text{driller}$ & $\text{duck}$ & $\text{eggbox*}$ & $\text{glue*}$ & $\text{holep}$ & $\text{iron}$ & $\text{lamp}$ & $\text{phone}$ & $\text{avg}$ \\
\cmidrule{2-5}
 & GS-ICP & Camera refine & Object Refine & GS-light &  &  &  &  &  &  &  &  &  &  &  &  &  &  \\
\midrule
$\bullet$ &  & & &  & 19.7 & 92.7 & 14.1 & 62.1 & 47.7 & 90.8 & 38.1 & 1.7 & 58.7 & 51.3 & 62.7 & 83.7 & 52.4 & 52.0\\
\midrule
$\bullet$ & $\bullet$ & & &  & 58.9 & 99.9 & 99.2 & 89.7 & 87.4 & 96.4 & 83.4 & 98.4 & 99.0 & 81.4 & 97.1 & 86.3 & 93.9 & 90.1 \\
$\bullet$ & & $\bullet$ & &  & 29.6 & 91.0 & 84.3 & 49.7 & 37.7 & 62.6 & 40.4 & 2.6 & 71.6 &  32.3 & 80.4 & 78.4 & 25.8 & 52.8 \\
$\bullet$ & & & $\bullet$ &  & 27.9 & 91.0 & 82.9 & 36.2 & 35.8 & 60.2 & 37.3 & 2.3 & 72.6 & 31.7 & 77.5 & 77.8 & 23.8 & 50.5 \\
$\bullet$ & & & & $\bullet$ & 30.5 & 91.0 & 84.4 & 47.5 & 38.1 & 63.4 & 40.6 & 2.6 & 72.9 & 33.7 & 81.2 & 78.4 & 26.1 & 53.1 \\
\midrule
$\bullet$ & $\bullet$ & $\bullet$ & & & 72.4 & 98.5 & 98.9 & 80.9 & 87.1 & 95.9 & 89.5 & 100.0 & 99.9 & 85.2 & 96.9 & 88.9 & 92.9 & 91.3\\
$\bullet$ & $\bullet$ & & $\bullet$ &  & 86.3 & 93.4 & 99.4 & 94.1 & 91.7 & 95.3 & 94.5 & 100.0 & 100.0 & 95.7 & 98.8 & 93.3 & 95.4 & 95.2 \\
$\bullet$ & $\bullet$ & & & $\bullet$ & 86.5  & 93.5 & 99.4 & 94.2 & 91.7 & 95.2 & 94.5 & 100.0 & 100.0 & 95.7 & 98.7 & 94.9 & 95.3 & 95.4 \\
$\bullet$ &  & $\bullet$ & $\bullet$ & & 29.5 & 91.0 & 84.3 & 49.5 & 37.7 & 62.6 & 40.4 & 2.6 & 71.6 & 32.4 & 80.4 &  78.4 & 25.8 & 52.8 \\
$\bullet$ &  & $\bullet$ & & $\bullet$ &  30.3 & 91.0 & 84.4 & 49.7 & 38.2 & 63.4 & 40.6 & 2.8 & 72.9& 33.7 & 81.2 & 78.3 & 25.9 & 53.3 \\
$\bullet$ &  &  & $\bullet$ & $\bullet$ & 30.5 & 91.0 & 84.4 & 49.8 & 38.2 & 66.3 & 40.6 & 2.6 & 72.9 & 33.7 & 81.1 & 78.4 & 25.9 & 53.5 \\
\midrule
$\bullet$ & $\bullet$ & $\bullet$ & $\bullet$ &  & 94.8 & 99.8 & 98.7 & 98.5 & 99.2 & 98.2 & 98.5 & 99.8 & 99.7 & 95.0 & 98.7 & 97.0 & 98.0 & 98.1 \\
$\bullet$ & $\bullet$ & $\bullet$ & & $\bullet$& 95.0 & 99.9 & 99.7 & 96.1 & 99.1 & 97.7 & 97.0 & 99.8 & 100.0 & 98.4 & 98.8 & 98.5 & 97.5 & 98.3 \\
$\bullet$ & $\bullet$ &  & $\bullet$ & $\bullet$& 94.1 & 99.9 & 99.4 & 94.2 & 98.4 & 95.3 & 97.6 & 100.0 & 100.0 & 95.7  & 98.8 & 97.2 & 95.4 & 97.4  \\
\midrule
$\bullet$ & $\bullet$ & $\bullet$ & $\bullet$ &$\bullet$ & \textbf{98.8} & \textbf{100.0} & \textbf{100.0} & \textbf{99.8} & \textbf{99.8} & \textbf{99.9} & \textbf{98.9} & \textbf{100.0} & \textbf{100.0} & \textbf{99.9} & \textbf{99.9} & \textbf{99.8} & \textbf{99.8} & \textbf{99.8} \\
\bottomrule
\end{tabular}

\vspace{2mm}
\begin{minipage}{0.95\linewidth}
\footnotesize
\textbf{*} The results marked with * indicate symmetric objects.
\end{minipage}
\end{table*}

In order to evaluate the effectiveness of the proposed model, this section conducts a comparative analysis of its performance against a range of state-of-the-art deep-learning 6D pose estimation models.

\subsection{Experimental Dataset and Settings}

We select three publicly accessible datasets for 6D pose estimation: T-LESS,  LineMod-Occlusion (LMO) and Linemod (LM). The objects in these datasets are textureless, and there exist certain variations in illumination during the image acquisition process.

\noindent \textbf{Datasets} : 
The T-LESS dataset contains 30 industrial objects without significant texture, distinctive color, and reflective properties. Each object is captured with a dense sampling of viewpoints against a black background using a Primesense Carmine 1.09RGBD camera. The LM dataset was collected in chaotic environments with significant illumination variations. It consists of 13 sequences with pose annotation for 13 objects, each with about 1200 images. We follow \cite{han2024pckrf,wang2019densefusion} to use about 15\% of the RGBD images for training, the remaining for testing. Notably, we do not use the synthetic images for expanding the dataset. The LMO dataset extends the LineMod dataset by incorporating occlusions, which contains 8 heavily occluded objects. Follow the previous work, We use it only for testing.

\noindent \textbf{Evaluation Metics}: Following prior works, we report our results using standard metrics: VSD for the T-LESS dataset, and ADD(-S) metrics with the 10\% diameter threshold for the LM and LMO datasets.

\noindent \textbf{Implementation Details}: The method in this paper is supported by the Pytorch framework and executed on a NVIDIA GeForce RTX 3090 GPU. The coarse pose estimation network uses SGD as the optimizer for the model, with a learning rate of 0.001 and a step size of 0.0005. The training is conducted for 300 epochs. In the precise pose estimation network, the learning rate is set to 0.001, with a total of 175 iterations. 

\subsection{Evaluation on Two Benchmark Datasets}

1) Evaluation on the T-LESS Dataset: TABLE \ref{tab:T-LESS} demonstrates our results on the T-LESS dataset, GS2POSE can achieve a high success rate of 74.7\%, which is 1.4\% higher than the state of art method. Compared with the SS-POSE algorithm, which also adopts ICP for refinement in its methodology, our proposed method achieves an improvement of over 6.9\%. This result demonstrates the effectiveness of the 3DGS-based iterative optimization algorithm in handling textureless objects.

2) Evaluation on the LineMod-Occlusion Dataset: We conducted experiments on LMO datasets to verify the performance of our method under textureless and occlusion challange. TABLE \ref{tab:LineMOD-Occlusion} demonstrates our results. Compared to state-of-the-art methods, we achieve 72.4\% on ADD(S)\textless0.1d, which are 2.8\% higher than the state-of-the-art methods. At the same time, our model demonstrates a significant improvement in accuracy on small objects such as ape, cat, and duck. This is primarily because, under occlusion, the feature information on small object surfaces is more sparse, making it more challenging to establish feature consistence. Our model, however, relies more on contour and color information of the objects, and adopts a multi-stage pose refinement strategy, thus alleviating this issue.

3) Evaluation on the LineMod Dataset: Fig. \ref{fig:coarse_result} presents the visualization results of NOCS images generated by Pose-Unet on the LM dataset. A comparison between the NOCS images generated by Pose-Unet and the ground-truth shows that their color distributions are almost identical. This demonstrates that Pose-Unet is capable of accurately generating NOCS reference images from novel viewpoints. TABLE \ref{tab:LineMOD} shows our final results. For nearly all objects, our model achieves close-to-best pose estimation performance. Specially, the training of the GS2POSE model does not require the synthetic images. The use of synthetic images can effectively reduce overfitting during training, thereby improving the final pose estimation accuracy \cite{li2023nerf}. In our network, the GS-refiner model is a fully training-free network capable of correcting poses, thereby effectively alleviating the overfitting issue of the coarse pose estimation network. 

More stringent evaluation metrics R\textless5° and R\textless5°, t\textless1cm were employed to further assess the performance of the model. The experimental results are presented in TABLE \ref{tab:LM2}. The experimental results demonstrate that our method outperforms state-of-the-art methods on both R\textless5°and R\textless5°, t\textless1cm, with a lead of 2.5\% on the mean standard. Meanwhile, our model achieves an improvement of 3.6\% in R\textless5°. This is due to the presence of numerous mismatches when predicting poses through feature consistency association in the absence of texture. In contrast, GS2POSE utilizes a reprojection iterative correction method, which places greater emphasis on the degree of contour fitting, thereby enhancing the accuracy of pose prediction.


Meanwhile, we statistically analyzed the angular prediction error distributions of both Coarse Pose Estimator and the Pose Refiner in Fig. \ref{fig:fenbu}. It can be observed that Coarse Pose Estimator constrains almost all angular errors within $15^\circ$, with over $70\%$ of the errors falling within $5^\circ$. In comparison, Pose Refiner further improves the accuracy, with more than $90\%$ of the errors within $5^\circ$.

\subsection{Ablation Study}

A test image was selected to demonstrate the process of GS2POSE, as illustrated in Fig. \ref{fig:refine_liucheng}. Subsequently, ablation experiments were conducted on the proposed module using the LineMod dataset. The experimental results are shown in TABLE \ref{tab:ablation}, which indicate that using the Pose-Unet model alone for pose estimation achieves only 52\% accuracy. However, when the GS-ICP module for precise pose estimation is incorporated, the pose prediction accuracy improves by 38.1\%. This suggests that the inaccuracy in the coarse pose estimation is primarily due to errors in depth information prediction. Through the comparative analysis of the second and third group of ablation experiments, it was observed that when the GS-ICP module is not employed to initially correct the depth values, the Camera refine, Object refine and GS-light modules often fail to effectively optimize the pose through iterative refinement. This is attributed to the fact that, in the presence of significant depth errors, the 3DGS struggles to identify the correct gradient descent direction for pose optimization. After incorporating the GS-ICP module to correct the depth errors, the addition of the Object refine module yields a more pronounced improvement in accuracy compared to the Camera refine module, indicating that after correcting the depth information with GS-ICP, the remaining pose inaccuracies are mainly caused by angular errors. When the GS-light module is incorporated, the pose accuracy improves by 1.7\%. This indicates that the color adaptation module effectively accommodates illumination variations in the scene, thereby enhancing the pose prediction accuracy.

\begin{table}[!t]
\centering
\setlength{\tabcolsep}{1mm} 
\small 
\begin{tabular}{c|*{4}{c}}
\toprule
Method & $R{<}5^\circ$ & $R{<}5^\circ, t{<}1\text{cm}$ & $\text{ADD}_{10}$ & Mean \\
\midrule
DenseFusion & 80.8 & 79.6 & 94.3 & 84.9 \\
PVN3D & 64.0 & 63.5 & 99.4 & 75.6 \\
FFB6D & \underline{87.9} & \underline{87.3} & \underline{99.7} & \underline{91.6} \\
\midrule
GS2Pose & \textbf{91.5(3.6$\uparrow$)} & \textbf{91.1(3.8$\uparrow$)} & \textbf{99.8(0.1$\uparrow$)} & \textbf{94.1(2.5$\uparrow$)}  \\
\bottomrule
\end{tabular}

\caption{More stringent evaluation metrics on the LM Dataset. Underlined and bold values represent the second-best and best results, respectively.}
\label{tab:LM2}
\end{table}

\section{Conclusion}

This paper presents GS2Pose, a novel two-stage 6D pose estimation algorithm based on the 3D Gaussian Splatting model. GS2POSE can achieve CAD-model-free operation while demonstrate strong applicability to textureless objects.

1) To enhance the robustness in estimating the pose of textureless objects, we propose an innovative Pose Refinement Network, which directly uses images to refine the object translation and rotation parameters. This network relies on capturing the contour and color features of the objects to refine their poses, instead of the texture features.

2) To address the challenge of depth blur, a ray projection-based point cloud registration method has been proposed. It can achieve accurate registration between the point clouds generated by 3DGS and RGBD images respectively, thereby resolving depth prediction errors.

3) To improve adaptability to variations in illumination, we propose an environment-adaptive strategy that can adjust the 3DGS color parameters according to illumination changes.

Compared to previous models. GS2POSE demonstrates accuracy improvements of 1.4\%, 2.8\%, 2.5\% on T-LESS, LineMod-Occlusion and LineMod datasets, respectively.

However, GS2POSE still has certain limitations. Currently, it can only address instance-level pose estimation and does not support one-shot pose estimation. Additionally, GS2POSE predicts poses through iterative refinement, which results in relatively low speed performance. In future work, we aim to enhance the model's generalization capability to unseen objects while simultaneously accelerating the convergence speed of the iterative process.

\section*{Acknowledgements}
This work was supported in part by the National Natural Science Foundation of China (Grants Nos. 62271016, 92148206) and the Beijing Natural Science Foundation under Grant 4222007.

\bibliographystyle{IEEEtran}
\bibliography{reference}

\end{document}